\RequirePackage{fix-cm}
\documentclass{article}

\usepackage{graphicx}

\usepackage{url}

\usepackage{booktabs}

\usepackage{color}
\usepackage{amsmath}
\usepackage{algorithmic}
\usepackage{pgf}
\usepackage{tikz}
\usetikzlibrary{arrows,shapes,automata,backgrounds,positioning}
\usetikzlibrary{matrix}
\tikzset{ 
    table/.style={
        matrix of nodes,
        row sep=-\pgflinewidth,
        column sep=-\pgflinewidth,
        nodes={
            rectangle,
            draw=black,
            align=center
        },
        minimum height=1.5em,
        text depth=0.5ex,
        text height=2ex,
        nodes in empty cells,
        every even row/.style={
            nodes={fill=gray!20}
        },
    }
}
\usepackage[caption=false]{subfig}
\usepackage{multirow}
\usepackage{url}
\usepackage{multirow}
\usepackage{amsfonts}

\usepackage{mathtools}
\usepackage{bm}
\usepackage{cite}

\usepackage{algorithmic}
\newcommand{\humanloss}[0]{0.036981}
\newcommand{\bestdevloss}[0]{0.036821}
\newcommand{\bestdevlossTEST}[0]{0.035192}
\newcommand{\trainingSessions}[0]{341} 
\definecolor{verylightgray}{RGB}{220,220,220}
\definecolor{lightblue}{RGB}{190,190,250}
\definecolor{lightgreen}{RGB}{160,255,160}

\definecolor{myblue}{RGB}{0,10,120}

\newcommand{\NUMBEROFSAMPLES}[1]{ 13406 }

\usepackage[normalem]{ulem}

\usepackage{authblk} 

\begin{document}

\title{A Graph Neural Network to Model Disruption in Human-Aware Robot Navigation
\thanks{This paper is an extension of~\cite{manso2020graph}. New contributions include considering the speed and previous poses of the robot and the people around it. It also provides a new dataset.}
\thanks{Submitted to Multimedia Tools and Applications, Springer.}
}

% \author{P. Bachiller$^1$ \and D. \nobreakdash{Rodriguez\nobreakdash-Criado}$^2$ \and R. R. Jorvekar$^3$ \and P. Bustos$^1$ \and D. R. Faria$^2$ \and L. J. Manso$^2$}
% 
% \date{%
%     $^1$Robotics  and  Artificial  Vision  Laboratory, University of Extremadura, Extremadura, Spain \texttt{\{pilarb,pbustos\}@unex.es}
%               \\%
%     $^2$College  of Engineering and Physical Sciences, Aston University, B4 7ET Birmingham, UK \texttt{190229717,d.faria,l.manso\}@aston.ac.uk}\\[2ex]%
%     $^3$Department of Computer Engineering, Pune Institute of Computer Technology, India\\[2ex]%
%     \today
% }

\author[1]{P. Bachiller}
\author[2]{D. Rodriguez-Criado}
\author[3]{R. R. Jorvekar}
\author[1]{P. Bustos}
\author[2]{D. R. Faria}
\author[2]{L. J. Manso}
\affil[1]{\normalsize Robotics  and  Artificial  Vision  Laboratory, University of Extremadura, Extremadura, Spain \texttt{\{pilarb,pbustos\}@unex.es}}
\affil[2]{\normalsize College  of Engineering and Physical Sciences, Aston University, B4 7ET Birmingham, UK \texttt{\{190229717,d.faria,l.manso\}@aston.ac.uk}}
\affil[3]{\normalsize Department of Computer Engineering, Pune Institute of Computer Technology, India}

\date{January 31, 2021}

\maketitle

\begin{abstract}
Autonomous navigation is a key skill for assistive and service robots.
To be successful, robots have to minimise the disruption caused to humans while moving.
This implies predicting how people will move and complying with social conventions.
Avoiding disrupting personal spaces, people's paths and interactions are examples of these social conventions.
This paper leverages Graph Neural Networks to model robot disruption considering the movement of the humans and the robot so that the model built can be used by path planning algorithms.
Along with the model, this paper presents an evolution of the dataset SocNav1~\cite{manso2020socnav} which considers the movement of the robot and the humans, and an updated scenario-to-graph transformation which is tested using different Graph Neural Network blocks.
The model trained achieves close-to-human performance in the dataset.
In addition to its accuracy, the main advantage of the approach is its scalability in terms of the number of social factors that can be considered in comparison with handcrafted models.
The dataset and the model are available in a public repository\footnote{https://github.com/gnns4hri/sngnnv2}.
\end{abstract}

\section{Introduction}
\label{sec:Intro}
Human-aware robot navigation deals with the challenge of endowing mobile social robots with the capability of considering the emotions and safety of people nearby while moving around their surroundings.
There is a wide range of works studying human-aware navigation from considerably diverse perspectives.
Pioneering works such as~\cite{Pacchierotti2005} started taking into account the personal spaces of the people surrounding the robots, often referred to as proxemics.
Semantic properties were also considered in~\cite{Cosley2009}.
In addition to proxemics, human motion patterns were analysed in~\cite{Hansen2009} to estimate whether humans are willing to interact with a robot.
Although not directly applied to navigation, the relationships between humans and objects were used in the context of ambient intelligence in~\cite{Bhatt2010}.
Proxemics and object affordances were jointly considered in~\cite{Vega2019} for navigation purposes.
Two extensive surveys on human-aware navigation can be found in~\cite{Rios-Martinez2015} and~\cite{Charalampous2017}.

\par

Despite the previously mentioned approaches being built on well-studied psychological models, they have limitations.
Considering new factors programmatically (\textit{i.e.}, writing additional code) involves a potentially high number of coding hours, makes systems more complex, and increases the chances of including bugs.
Additionally, with every new aspect to be considered for navigation, the decisions made become less \textit{explainable}, which is precisely one of the main advantages of handcrafted approaches over data-driven ones.
In addition to the mentioned model scalability and explainability issues, handcrafted approaches have the intrinsic and rather obvious limitation that they only account for what the model explicitly considers.
Given that these models are manually written by humans, they cannot account for aspects that the designers are not aware of.
% ****************************************************************************
% * THIS IS SOMETIMES REFERRED TO AS THE: "KNOWLEDGE ACQUISITION BOTTLENECK" *
% ****************************************************************************
% We can mention it in the next round.
% 
\par

Approaches leveraging machine learning have also been published.
The parameters of a social force model~\cite{Helbing1995} are learned in~\cite{Ferrer2013} and~\cite{Patompak2019} to navigate in human-populated environments.
Inverse reinforcement learning is used in~\cite{Ramon-Vigo2014} and~\cite{vasquez2014inverse} to plan navigation routes based on a list of humans in a radius.
Social norms are implemented using deep reinforcement learning in~\cite{Chen2017}, again, considering a set of humans.
An approach modelling crowd-robot interaction and navigation control is presented in~\cite{Chen2019}.
It features a two-module architecture where single interactions are modelled and then aggregated.
Although its authors reported good qualitative results, the approach does not contemplate integrating additional information (\textit{e.g.}, relations between humans and objects, structure and size of the room).
The work in~\cite{martins2019clusternav} tackles the same problem using Gaussian Mixture Models.
It has the advantage of requiring less training data, but the approach is also limited in terms of the input information it can process.

\par

All the previous works and many others not mentioned have achieved outstanding results.
Some model-based approaches such as~\cite{Cosley2009} or~\cite{Vega2019} can leverage structured information to take into account space affordances.
Still, the data considered to make such decisions are often handcrafted features based on an arbitrary subset of the data that a robot can potentially work with.
There are many reasons motivating to seek learning-based approaches not limited to a selection of handcrafted features.
Their design is time-consuming and often requires a deep understanding of the particular domain (see discussion in~\cite{Lecun2015}).
Additionally, there is generally no guarantee that a particular hand-engineered set of features is close to being the best possible one.
On the other hand, most end-to-end deep learning approaches have important limitations too.
They require a large amount of data and computational resources that are often scarce and expensive, and they are hard to explain and manually fine-tune.
Somewhere in the middle of the spectrum, we have proposals advocating not to choose between hand-engineered features or end-to-end learning.
In particular, \cite{Battaglia2018} proposes Graph Neural Networks (GNNs) as a means to perform learning that allows combining raw data with hand-engineered features, and most importantly, to learn from structured information.
The relational inductive bias of GNNs is specially well-suited to learn about structured data and the relations between different types of entities, often requiring less training data than other approaches.
In this line, we argue that using GNNs for human-aware navigation reduces the time and effort required to integrate new social cues.

\par

In this work, we trained different GNN models to estimate people's comfort given a scenario and its previous states.
The state of a scenario includes objects, walls, the robot, and humans who can be interacting with other humans or objects.
For moving entities (\textit{i.e.}, humans and the robot) the network also considers not only their pose but also their linear and angular velocities.
GNNs are proposed because the information that social robots can work with is not just a map and a list of people, but a more sophisticated data structure where the entities represented can have different relations among them.
For example, social robots could potentially have information about who a human is talking to, where people are looking at, who is friends with whom, or who is the owner of an object in the scenario.
Regardless of how this information is acquired, it can be naturally represented using a graph, and GNNs are a particularly well-suited and scalable machine learning approach to work with these graphs.

\section{Graph Neural Networks}
\label{sec:GNN}
\subsection{Graph Neural Networks basics}
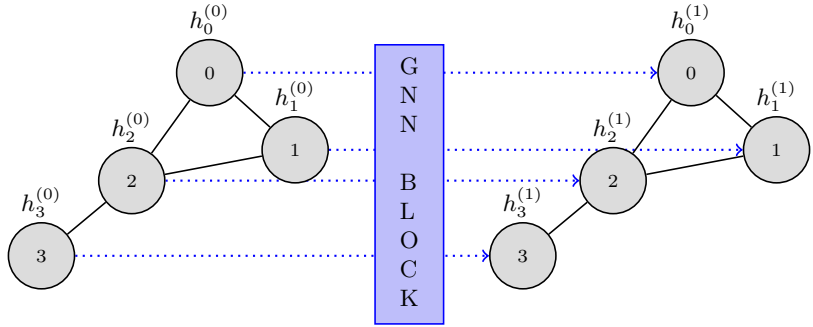
\begin{figure*}[ht]
\centering
\begin{tikzpicture}[font=\small,auto,semithick]
  \tikzstyle{every state}=[fill=verylightgray,draw=black,text=black,font=\scriptsize]
  \node[state](n0a) [] {$0$};
  \node[](n0at) [above=-0.6mm of n0a] {$h^{(0)}_0$};
  \node[state](n1a) [below right=3.9mm and 5mm of n0a] {$1$};
  \node[](n1at) [above=-0.6mm of n1a] {$h^{(0)}_1$};
  \node[state](n2a) [below left=8mm and 4mm of n0a] {$2$};
  \node[](n2at) [above=-0.6mm of n2a] {$h^{(0)}_2$};
  \node[state](n3a) [below left=18mm and 16mm of n0a] {$3$};
  \node[](n3at) [above=-0.6mm of n3a] {$h^{(0)}_3$};
  \node[state](n0b) [right=55mm of n0a] {$0$};
  \node[](n0bt) [above=-0.6mm of n0b] {$h^{(1)}_0$};
  \node[state](n1b) [below right=3.9mm and 5mm of n0b] {$1$};
  \node[](n1bt) [above=-0.6mm of n1b] {$h^{(1)}_1$};
  \node[state](n2b) [below left=8mm and 4mm of n0b] {$2$};
  \node[](n2bt) [above=-0.6mm of n2b] {$h^{(1)}_2$};
  \node[state](n3b) [below left=18mm and 16mm of n0b] {$3$};
  \node[](n3bt) [above=-0.6mm of n3b] {$h^{(1)}_3$};
  \begin{scope}
  \node[rectangle](XXX) [below right=-7mm and 18.7
  mm of n0a,color=blue,fill=lightblue,draw,rectangle] {\color{black}\begin{tabular}{l} G \\ N \\ N \\ \\ B \\ L \\ O \\ C \\ K \end{tabular}};
  \end{scope}

  \begin{scope}[on background layer]
  \begin{scope}[dotted, blue, thick]
    \draw[->](n0a) to[out=0,in=180](n0b);
    \draw[->](n1a) to[out=0,in=180](n1b);
    \draw[->](n2a) to[out=0,in=180](n2b);
    \draw[->](n3a) to[out=0,in=180](n3b);
  \end{scope}
  \end{scope}
  \path
        (n1a) edge node {} (n0a)
        (n2a) edge node {} (n0a)
        (n2a) edge node {} (n1a)
        (n3a) edge node {} (n2a)
        (n1b) edge node {} (n0b)
        (n2b) edge node {} (n0b)
        (n2b) edge node {} (n1b)
        (n3b) edge node {} (n2b);
\end{tikzpicture}
\caption{A basic GNN block/layer. GNN layers output updated versions of the input graph. These updated graphs have the same nodes and links, but the feature vectors of the nodes will generally differ in size and content depending on the feature vectors of their neighbours and their own vectors in the input graph. A GNN is usually composed of several stacked GNN layers. Higher level features are learnt in the deeper layers, so that the output of any of the nodes in the last layer can be used for classification or regression purposes.}
\label{fig:gnnlayer}
\end{figure*}

Graph Neural Networks (GNNs) are a family of machine learning approaches which extend neural networks to be able to take graph-structured data as input.
They can perform classifications and regressions on graphs, nodes, edges, as well as predicting links when working with partially observable phenomena.
Except for few exceptions (\textit{e.g.},~\cite{ying2018hierarchical}) GNNs are composed by similar stacked blocks (layers) operating on a graph whose structure remains static but the features associated to its nodes are updated in every layer of the network (see Fig.~\ref{fig:gnnlayer}).

\par

As a consequence, the features associated to the nodes of the graph in each layer become more abstract and are influenced by a wider context as layers go deeper.
The features in the nodes of the last layer are frequently used to perform the final classification or regression.

\par

The first published efforts on applying neural networks to graphs date back to works by A. Sperduti et al.~\cite{Sperduti1998}.
GNNs were further studied and formalised by M. Gori et al.~\cite{Gori2005} and F. Scarselli et al.~\cite{F.2009}.
However, it was with the appearance of Gated Graph Neural Networks~\cite{Li2015} and especially Graph Convolutional Networks (GCNs, \cite{Kipf2016a}) that GNNs gained traction.
A review and a unified notation for GNNs can be found in~\cite{Battaglia2018}.

\par
Graph Convolutional Networks (GCN~\cite{Kipf2016a}) is one of the most common GNN blocks.
Because of its simplicity, we build on the GCN block to provide the reader with an intuition of how GNNs work.
Following the notation proposed in~\cite{Battaglia2018}, GCN blocks operate over a graph $G=(V,E)$, where $V=\{v_i\}_{i=1:N^v}$ is a set of nodes, being $v_i$ the feature vector of node $i$ and $N^v$ the number of vertices in the graph.
$E=\{(s_k,r_k)\}_{k=1:N^e}$ is a set of edges where $s_k$ and $r_k$ are the source and destination indices of edge $k$ and $N^e$ is the number of edges in the graph.
Each GCN layer generates an updated representation $v_i'$ for each node $v_i$ using two functions:
$$\displaystyle\rho^{e \rightarrow v}(E_i) = \sum_{\{k:r_k=i\}}e_k,$$
$$\phi^{v}(\overline{e}_i, v_i)=NN_v([\overline{e}_i,v_i]).$$
For every node $v_i$, the first function ($\displaystyle\rho^{e \rightarrow v}(E_i)$) is used to aggregate the feature vectors of other nodes with an edge towards $v_i$ and generates a temporary aggregated feature $\overline{e}_i$ which is used by the second function:
$$\overline{e}_i = \displaystyle\rho^{e \rightarrow v}(E_i).$$
The function $\phi^{v}(\overline{e}_i, v_i)$ is then used to generate an updated $v_i'$ feature vector for each node $i$ from the aggregated feature vector $\overline{e}_i$ using a neural network (usually a multi-layer perceptron, but the framework does not make any assumption on this):
$$v_i' = \phi^{v}(\overline{e}_i, v_i).$$
Such a learnable function is generally the same for all the nodes.
By stacking several blocks where features are aggregated and updated, the feature vectors can carry information from nodes far away in the graph and convey higher level features that can be finally used for classification or regressions.

\par

Several means of improving GCNs have been proposed.
Relational Graph Convolutional Networks (R-GCNs~\cite{Schlichtkrull2018}) extend GCNs by considering different types of edges separately.
They are applied to vertex classification and link prediction in ~\cite{Schlichtkrull2018}.
Graph Attention Networks (GATs~\cite{Velickovic2018}) extend GCNs by adding self-attention mechanisms (see~\cite{vaswani2017attention}).
They are applied to vertex classification in~\cite{Velickovic2018}.
In Message Passing Graph Neural Networks (MPNNs~\cite{gilmer2017neural}), the messages which are aggregated are not only composed of node features but also edge features.
This allows MPNNs to account for both vertex and edge features.
For a more detailed review of GNNs and the generalised framework, please refer to~\cite{Battaglia2018}.

\subsection{Graph Neural Networks applied to human-aware navigation}
A number of recent machine learning-based approaches leveraging structured data for social navigation have been recently published.
A GNN model integrated with a Deep Reinforcement Learning (DRL) algorithm based in Monte Carlo Tree Search was presented in~\cite{Chen2019GNN}.
It utilises a graph-based model to detect the implicit relations between the humans in a room.
Interactions are useful to predict future human trajectories.
For instance, interacting pedestrians generally behave differently than those who do not interact.
This phenomena is also exploited in~\cite{Chen2020}, where a GCN-based DRL leverages the gaze of humans to estimate interactions and predict their trajectories.
These works consider human-robot and human-human relations but  disregard interactions with objects or obstacles that could be exploited.
Moreover, the DRL algorithms in~\cite{Chen2019GNN} and~\cite{Chen2020} use a simple handcrafted reward function based on the minimum distance between the robot and the humans that disregards any other information including the orientation and velocity of the humans or how densely populated the room is (distance restrictions are usually eased in crowded spaces).
Due to the variety of different scenarios and factors to consider, handcrafting a reward function that complies with social rules seems prohibitively complex and time-consuming.

\par

A model combining a Convolutional Neural Network (CNN) and a GNN to learn an action policy for multi-robot navigation is presented in~\cite{Li2019}.
The CNN extracts features from local observations of the environment, and the action policy for the robot swarm is computed from those features using GNN.
Although safety and collision avoidance are considered, the approach only considers humans as obstacles.

\par

Other works use GNNs for reasoning and perception in the domain of social navigation. 
A significant amount of them directs their focus to the prediction of pedestrians' paths as exemplified in works undertaken by~\cite{Vemula2017}, \cite{Huang2019} or~\cite{Haddad2020}.
The use of GNNs for these tasks allows extracting additional information from the crowd such as relations between people.
However, none of the previous works tackle the problem of modelling discomfort.

\par

GNNs have been used to model and estimate discomfort in our previous works, \cite{manso2020graph} and~\cite{Rodriguez-Criado2020}.
Both works generate discomfort estimations in a scale from 0 to 100 and consider human-human, human-robot and human-object interactions, as well as walls and other objects.
While~\cite{manso2020graph} generates a single value a given scenario, \cite{Rodriguez-Criado2020} generates a two-dimensional cost map using a combination of GNNs and CNNs, in that order.
The main limitation of these models is that the scenarios they consider are static (\textit{i.e.}, they disregard human and robot motion). 

\par

The work at hand follows a similar approach to~\cite{manso2020graph} with a number of enhancements.
Firstly, we consider two different scores to measure two aspects of social navigation (see section~\ref{socnav2}).
Secondly, the model is trained using dynamic scenes where humans and the robot move, which was the main limitation of~\cite{manso2020graph}.

\par

\section{SocNav2 dataset}\label{socnav2}
SocNav1~\cite{manso2020socnav}, was designed to learn and benchmark estimation functions for social navigation conventions.
\textit{SocNav2} -presented in this paper- has the same goal as its predecessor but unlike SocNav1, it considers the velocity and trajectory of the robots and the humans around them. 
As SocNav2, SocNav1 contains scenarios with a robot in a room, a number of objects and a number of people that can potentially be interacting with other objects or people.
In case any human\nobreakdash-human or human\nobreakdash-object interaction exists it is also noted in the scenarios.
Each sample in the dataset is given a score between $0$ and $100$, depending on the extent that the subjects consider that the robot is disturbing the people in the scenario. 
The main limitation of SocNav1 is that samples do not consider velocity information or the trajectory of the humans.
\par
SocNav2 overcomes such limitation and provides $\NUMBEROFSAMPLES{}$ scored samples of dynamic scene sequences. 
Each sample consists of $35$ ``snapshots'' of a scene of a room with a moving robot, objects and potentially moving humans, taken during a time interval of 4 seconds.
In SocNav2 the room also includes a landmark that constitutes a goal position to be reached by the robot.
\par
Each SocNav2 sample includes scores for two social navigation-related statements: ``\textit{the robot does not cause any disturbance to the humans in the room}'' (\textit{\textbf{Q1}}) and ``\textit{the robot is moving towards the goal efficiently, not causing any disturbance to the humans in the room}'' (\textit{\textbf{Q2}}). The scores range from $0$ to $100$, considering the following reference values:
\begin{itemize}
    \item 0: unacceptable
    \item 20: undesirable
    \item 40: acceptable
    \item 60: good
    \item 80: very good
    \item 100: perfect
\end{itemize}
\par
The scenarios compiled in SocNav2 have been generated using SONATA~\cite{baghel2020toolkit}.
SONATA is a toolkit built on top of PyRep~\cite{james2019pyrep} and CoppeliaSim~\cite{rohmer2013coppeliasim} designed to simulate human-populated navigation scenarios and to generate datasets.
It provides an API to generate random scenarios including humans, objects, interactions, the robot and its goals.
The walls delimiting a room are also randomly generated considering rectangular and L-shaped rooms.
Despite SONATA only provides simulated scenarios, the use of synthetic data is essential in the context of social navigation.
Firstly, because it would not be feasible to generate as many situations using only real-world data.
Secondly, because situations endangering humans' integrity, such as human-robot collisions, could not be generated in real scenarios.
\par
The movements of the robot were generated through two different strategies to increase the diversity of its behaviour.
The first strategy uses a machine learning model (see~\cite{baghel2020toolkit}) that outputs the control actions of the robot according to a graph representation of the scenario.
This model was trained using supervised learning (\textit{i.e.}, it only contains examples of appropriate behaviours), so it has unexpected behaviours in situations that would not usually happen when controlled by humans.
Nevertheless, for the creation of SocNav2, these behaviours allow to generate a wide variety of good and bad situations that would not have been obtained from random actions.
In addition to the samples where the movement of the robot was controlled by the machine learning approach, a second set of samples was generated using a joystick to control the robot manually.
This second set was created to include infrequent situations in the first set, such as the robot moving backwards to avoid getting blocked or stopping to let people pass.
\par
\begin{figure*}[ht] 
    \centering
    \subfloat[First second.]{\includegraphics[width=0.45\textwidth,keepaspectratio=true]{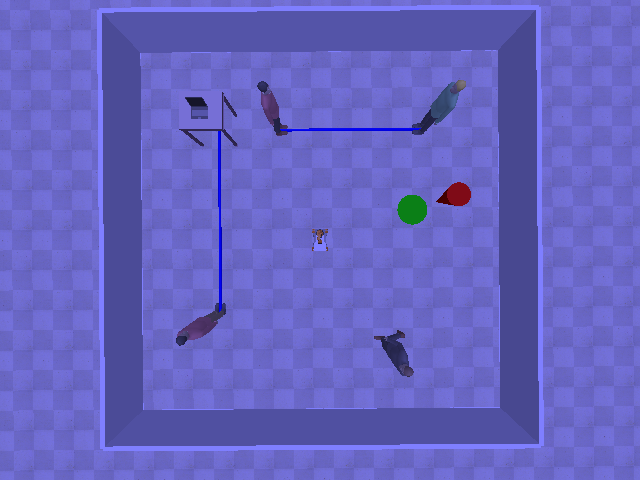} \label{fig:img_scn3_f0}}
    \hspace{0.4cm}
    \subfloat[Second second.]{\includegraphics[width=0.45\textwidth,keepaspectratio=true]{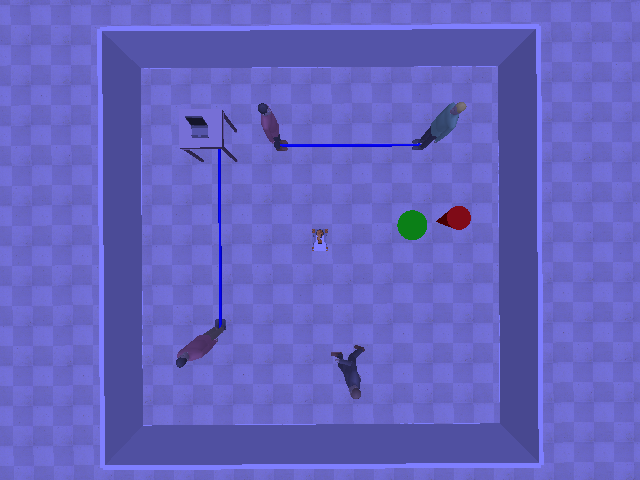} \label{fig:img_scn3_f1}} \\
    \subfloat[Third second.]{\includegraphics[width=0.45\textwidth,keepaspectratio=true]{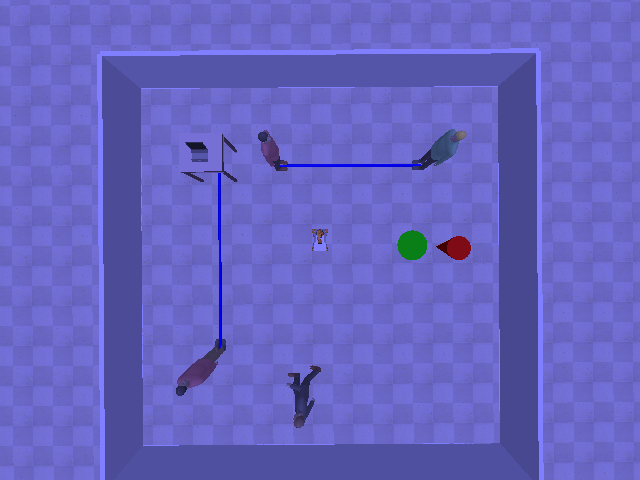} \label{fig:img_scn3_f2}} 
    \hspace{0.4cm}
    \subfloat[Fourth second]{\includegraphics[width=0.45\textwidth,keepaspectratio=true]{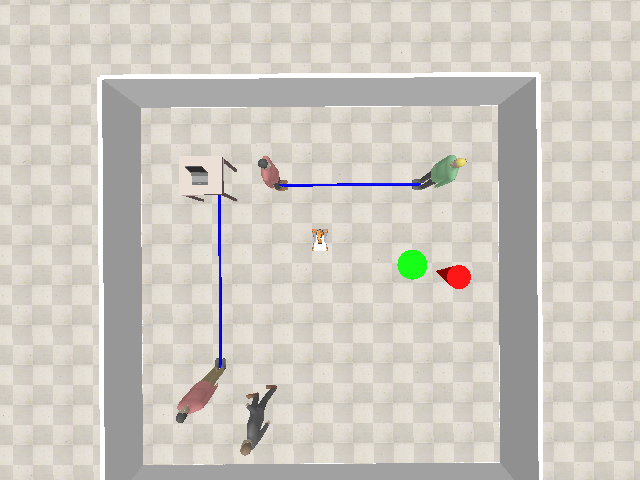} \label{fig:img_scn3_f3}}
    \caption{A SocNav2 sequence. The shaded images correspond to the first 3 seconds of the sequence, which are also shown to subjects to provide context. The last image, in Fig.~\ref{fig:img_scn3_f3}, corresponds to the second that the users score. During the whole sequence the robot is moving forwards.}
    \label{fig:scenario_3}
\end{figure*}

The subjects providing the scores for SocNav2 were shown sequences of 4 seconds, and were asked to give their answers for the behaviour of the robot in the last second.
During the three previous seconds, the video was shaded to make easier to know what time slice had to be evaluated (see~Fig.~\ref{fig:scenario_3}).
The geometrical and relational data of the sequences were stored in JSON files.
Subjects were asked to provide a score for \textit{Q1} and \textit{Q2} after watching the video (as many times as necessary) according to the aforementioned reference values.
Despite some guidelines were given, subjects were requested to feel free to express their opinions.
Some of the guidelines were the following:
\begin{itemize}
    \item The goal should be disregarded when answering Q1. It should only be considered when answering Q2.
    \item The closer the robot gets to people, the more it can be deemed disturbing.
    \item In small rooms with a high number of people, closer distances are acceptable in comparison to big rooms with fewer people.
    \item The robot is required not to collide with objects or walls. If it collides it should have a score of $0$.
\end{itemize}

Six subjects participated in the scoring of the dataset, producing $\NUMBEROFSAMPLES{}$  scored samples.
This initial set of samples has been extended using data augmentation.
Specifically, each scenario has been mirrored in the vertical axis assuming the same scores as in the original scenario.
In addition, each normal and mirrored scenario has been rotated $180^\circ$, changing also the sign of the advance speed of the robot. 
This extension assumes that the human discomfort does not change whether the robot is moving forwards or backwards.
As a result of this data augmentation process, the final dataset is composed of $53600$ samples.
\par
In order to analyse the consistency of the scoring of the dataset, the inter-rater and intra-rater agreements have been computed for 4 subjects using the linearly weighted kappa coefficient~\cite{Cohen1968}. 
For the inter-rater consistency, common samples scored by each pair of subjects were considered. 
The minimum number of common samples for which this coefficient was obtained is $609$.
For measuring the intra-rater reliability, each user scored 200 duplicate samples.
Tables~\ref{tab:inter_intra_Q1} and~\ref{tab:inter_intra_Q2} show the inter-rater and intra-rater consistency for the scores of Q1 and Q2, respectively (intra-rater in the diagonal cells, inter-rater in the remaining cells).
\par
As shown in Table~\ref{tab:inter_intra_Q1}, the intra-rater consistency for Q1 is ``\textit{almost perfect}'' in the scale defined in~\cite{Landis77}.
The inter-rater agreement for Q1 can be considered substantial in most of the cases.
Only subjects 1 and 4 have a low agreement, but they fall in the high \textit{moderate} bracket.
Table~\ref{tab:inter_intra_Q2} shows that the consistency for Q2 is generally lower than for Q1.
This reduction can be due to the very nature of the question, since subjects may broadly differ about how the robot should move to \textit{efficiently} reach the goal position.
Nevertheless, the inter-rater and intra-rater consistency is still \textit{substantial} excepting for subjects 1 and 4, which is high \textit{moderate}.

\begin{table}[ht]
\caption{Inter-rater and intra-rater consistency of four subjects for Q1.}
\centering
%% \tablesize{} %% You can specify the fontsize here, e.g., \tablesize{\footnotesize}. If commented out \small will be used.
\begin{tabular}{ccccc}
\toprule
\textbf{\hfill}	& \textbf{Subject1} & \textbf{Subject2} & \textbf{Subject3} & \textbf{Subject4} \\
\midrule
\textbf{Subject1} & 0.83 & 0.75 & 0.80 & 0.56\\
\midrule
\textbf{Subject2} & 0.75 & 0.88 & 0.85 & 0.63\\
\midrule
\textbf{Subject3} & 0.80 & 0.85 & 0.88 & 0.62\\
\midrule
\textbf{Subject4} & 0.56 & 0.63 & 0.62 & 0.81\\
\bottomrule
\end{tabular}
\label{tab:inter_intra_Q1}
\end{table}

\begin{table}[ht]
\caption{Inter-rater and intra-rater consistency of four subjects for Q2.}
\centering
%% \tablesize{} %% You can specify the fontsize here, e.g., \tablesize{\footnotesize}. If commented out \small will be used.
\begin{tabular}{ccccc}
\toprule
\textbf{\hfill}	& \textbf{Subject1} & \textbf{Subject2} & \textbf{Subject3} & \textbf{Subject4} \\
\midrule
\textbf{Subject1} & 0.74 & 0.68 & 0.72 & 0.57\\
\midrule
\textbf{Subject2} & 0.68 & 0.71 & 0.74 & 0.63\\
\midrule
\textbf{Subject3} & 0.72 & 0.74 & 0.76 & 0.64\\
\midrule
\textbf{Subject4} & 0.57 & 0.63 & 0.64 & 0.73\\
\bottomrule
\end{tabular}
\label{tab:inter_intra_Q2}
\end{table}

\par

\section{Scenario to graph transformation}\label{problem}
This paper follows the strategy developed in~\cite{manso2020graph} and includes a number of modifications to account for velocity and trajectory information.
To leverage the properties of GNNs (see section~\ref{sec:GNN}) the input data from \textit{SocNav2} has to be transformed into a graph.
This section describes the scenario\nobreakdash-to\nobreakdash-graph transformation process.

\subsection{Graph structure}
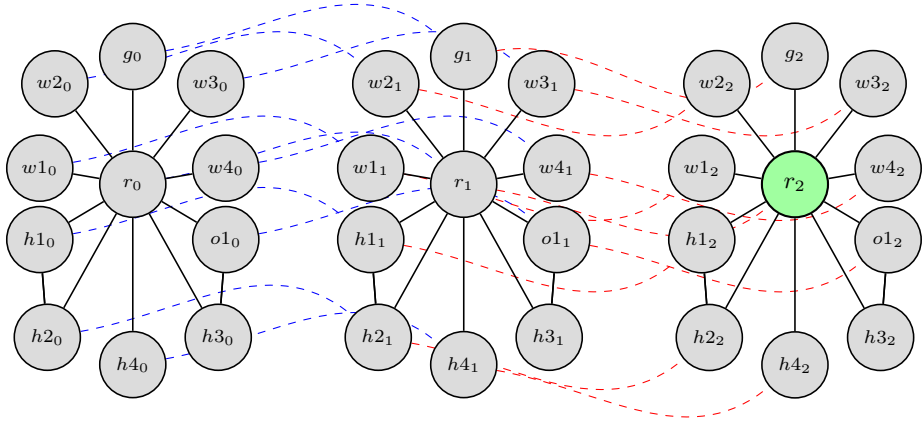
\begin{figure*}[t]
\centering
\begin{tikzpicture}[font=\small,auto,semithick]
  \tikzstyle{every state}=[fill=verylightgray,draw=black,text=black,font=\scriptsize]
  \node[state](room1) [] {$r_1$};
  \node[state](room0) [left=35mm of room1] {$r_0$};
  \begin{scope} [thick]
    \tikzstyle{every state}=[fill=lightgreen,text=black, thick]
    \node[state](room2) [right=35mm of room1] {$r_2$};
  \end{scope}
  \node[state](w10)   [below left=-8.5mm and 6mm of room0] {$w1_0$};
  \node[state](g0)   [above = 8mm of room0] {$g_0$};
  \node[state](w20)   [above left=7mm and 4mm of room0]  {$w2_0$};
  \node[state](w30)   [above right=7mm and 4mm of room0] {$w3_0$};
  \node[state](w40)   [below right=-8.5mm and 6mm of room0] {$w4_0$};
  \node[state](h10)   [below left=1mm and 6mm of room0]    {$h1_0$};
  \node[state](h20)   [below left=14mm and 5mm of room0]    {$h2_0$};
  \node[state](h30)   [below right=14mm and 5mm of room0]    {$h3_0$};
  \node[state](h40)   [below      =15mm   of room0]    {$h4_0$};
  \node[state](o10)   [below right=1mm and 6mm of room0]    {$o1_0$};
  \node[state](w11)   [below left=-8.5mm and 6mm of room1] {$w1_1$};
  \node[state](g1)   [above = 8mm of room1] {$g_1$};
  \node[state](w21)   [above left=7mm and 4mm of room1]  {$w2_1$};
  \node[state](w31)   [above right=7mm and 4mm of room1] {$w3_1$};
  \node[state](w41)   [below right=-8.5mm and 6mm of room1] {$w4_1$};
  \node[state](h11)   [below left=1mm and 6mm of room1]    {$h1_1$};
  \node[state](h21)   [below left=14mm and 5mm of room1]    {$h2_1$};
  \node[state](h31)   [below right=14mm and 5mm of room1]    {$h3_1$};
  \node[state](h41)   [below      =15mm   of room1]    {$h4_1$};
  \node[state](o11)   [below right=1mm and 6mm of room1]    {$o1_1$};
  \node[state](w12)   [below left=-8.5mm and 6mm of room2] {$w1_2$};
  \node[state](g2)   [above = 8mm of room2] {$g_2$};
  \node[state](w22)   [above left=7mm and 4mm of room2]  {$w2_2$};
  \node[state](w32)   [above right=7mm and 4mm of room2] {$w3_2$};
  \node[state](w42)   [below right=-8.5mm and 6mm of room2] {$w4_2$};
  \node[state](h12)   [below left=1mm and 6mm of room2]    {$h1_{2}$};
  \node[state](h22)   [below left=14mm and 5mm of room2]    {$h2_{2}$};
  \node[state](h32)   [below right=14mm and 5mm of room2]    {$h3_{2}$};
  \node[state](h42)   [below      =15mm   of room2]    {$h4_{2}$};
  \node[state](o12)   [below right=1mm and 6mm of room2]    {$o1_{2}$};
  \begin{scope}[on background layer]
  \begin{scope}   [dashed,thin, blue]
    \draw[-](room0) to[out=10,in=140](room1);
    \draw[-](g0) to[out=10,in=140](g1);
    \draw[-](w10) to[out=10,in=140](w11);
    \draw[-](w20) to[out=10,in=140](w21);
    \draw[-](w30) to[out=10,in=140](w31);
    \draw[-](w40) to[out=10,in=140](w41);
    \draw[-](h10) to[out=10,in=140](h11);
    \draw[-](h20) to[out=10,in=140](h21);
    \draw[-](h40) to[out=10,in=140](h41);
    \draw[-](o10) to[out=10,in=140](o11);
  \end{scope}
  \begin{scope}   [dashed,thin,red]
    \draw[-](room1) to[out=-10,in=-140](room2);
    \draw[-](g1) to[out=10,in=-140](g2);
    \draw[-](w11) to[out=-10,in=-140](w12);
    \draw[-](w21) to[out=-10,in=-140](w22);
    \draw[-](w31) to[out=-10,in=-140](w32);
    \draw[-](w41) to[out=-10,in=-140](w42);
    \draw[-](h11) to[out=-10,in=-140](h12);
    \draw[-](h21) to[out=-10,in=-140](h22);
    \draw[-](h41) to[out=-10,in=-140](h42);
    \draw[-](o11) to[out=-10,in=-140](o12);
  \end{scope}
  \end{scope}
  \path
        (w10) edge              node {} (room0)
        (g0) edge               node {} (room0)
        (w20) edge              node {} (room0)
        (w30) edge              node {} (room0)
        (w40) edge              node {} (room0)
        (h10) edge              node {} (room0)
        (h20) edge              node {} (room0)
        (h30) edge              node {} (room0)
        (h40) edge              node {} (room0)
        (o10) edge              node {} (room0)
        (h10) edge              node {} (h20)
        (h20) edge              node {} (h10)
        (h30) edge              node {} (o10)
        (o10) edge              node {} (h30)
        (w11) edge              node {} (room1)
        (g1) edge               node {} (room1)
        (w21) edge              node {} (room1)
        (w31) edge              node {} (room1)
        (w41) edge              node {} (room1)
        (h11) edge              node {} (room1)
        (h21) edge              node {} (room1)
        (h31) edge              node {} (room1)
        (h41) edge              node {} (room1)
        (o11) edge              node {} (room1)
        (h11) edge              node {} (h21)
        (h21) edge              node {} (h11)
        (h31) edge              node {} (o11)
        (o11) edge              node {} (h31)
        (w12) edge              node {} (room2)
        (g2) edge               node {} (room2)
        (w22) edge              node {} (room2)
        (w32) edge              node {} (room2)
        (w42) edge              node {} (room2)
        (h12) edge              node {} (room2)
        (h22) edge              node {} (room2)
        (h32) edge              node {} (room2)
        (h42) edge              node {} (room2)
        (o12) edge              node {} (room2)
        (h12) edge              node {} (h22)
        (h22) edge              node {} (h12)
        (h32) edge              node {} (o12)
        (o12) edge              node {} (h32);
\end{tikzpicture}
\caption{Example of how the scenario-to-graph transformation works, based on the scenario depicted in Fig.\ref{fig:scenario_3}}
\label{fig:graphs}
\end{figure*}

The graphs inputted to the GNN models are composed of a sequence of 3 sub-graphs for 3 \textit{snapshots} of the videos shown to the subjects.
Each sub-graph (\textit{frame graph}) is separated by 1 second, being the last one the graph which the users scored.
The graph creation process has two steps.
First, each \textit{snapshot} is transformed into a separate \textit{frame graph}.
Once the 3 frame graphs in the sequence have been generated, they are merged into a single graph representing the sequence (see Fig.~\ref{fig:graphs}).
This temporal connection is done with an edge linking the node in each frame graph with the same node in the next frame graph.
\par
The nodes in the graphs have five types:
\begin{itemize}
 \item \textbf{room:} There is one room node per frame graph. It acts as a global node~\cite{Battaglia2018} and it is connected to any other node of the graph for that frame. Using a global node favours communication across the graph and reduces the number of layers required.
 \item \textbf{wall:} A node for each of the segments defining the room. 
 \item \textbf{goal:} Used to represent the position that the robot must reach.
 \item \textbf{object:} A node for each object in the scenario.
 \item \textbf{human:} A node for each human. Humans might be interacting with objects or other humans.
\end{itemize}

\par
There is no node explicitly representing the robot because all node features are in the reference frame of the robot (further explained in section~\ref{sec:features}).
For every human engaging in interactions, two new edges are added between the human and the entity (human or object) they interact with, one in each direction.
The graphs also include self-edges for all nodes, and the room node is connected in both directions to the rest of the nodes in the graph.
As an example, Fig.~\ref{fig:scenario_3} depicts four frames of a sequence where four humans are in a room with several objects.
Two of the humans are interacting with each other, another human is interacting with an object, and the remaining human is moving without interacting with other human or object.
Fig.~\ref{fig:graphs} shows the structure of the resulting graph.

\subsection{Node and edge features}\label{sec:features}
\textbf{Node} feature vectors are built by concatenating different sections.
The first section is a one-hot encoding for the type of node.
The remaining sections are type specific and are only filled if the node is of the corresponding type, filled with zeros otherwise.
The features used in the sections for \textit{human}, \textit{wall} and \textit{object} nodes are: position, distance to the robot, speed and orientation, all from the robot's frame of reference.
Position and distance are represented in decametres for normalization purposes.
Similarly, the orientation is split into sine and cosine, instead of including the angle itself.
For wall segments, the position is the centre of the segment and the orientation is the tangent.
Object sections also contain width and height features defining the object's bounding box.
The section corresponding to the \textit{room} symbol is composed of the number of humans in the room and the velocity command given to the robot.
Table~\ref{tbl:nodefeatures} depicts this layout.

\begin{table}[ht]
\centering
\begin{tabular}{|l|c|c|c|c|c|}
\hline
\textbf{n. one-hot} & \multicolumn{5}{c|}{5 elements (one per node type)}                                              \\ \hline
\textbf{f. one-hot} & \multicolumn{5}{c|}{3 elements (one per frame graph)}                                            \\ \hline
\textbf{room}      & \multicolumn{2}{c|}{number of humans} & adv. speed   & \multicolumn{2}{c|}{rot. speed} \\ \hline
\textbf{human}     & pos. (2D)      & speed (3D)       & orientation (2D)   & \multicolumn{2}{c|}{distance}   \\ \hline
\textbf{object}    & pos. (2D)      & speed (3D)       & orientation (2D)   & distance       & shape (2D)      \\ \hline
\textbf{wall}      & pos. (2D)      & \multicolumn{2}{c|}{orientation (2D)} & \multicolumn{2}{c|}{distance}   \\ \hline
\textbf{goal}      & \multicolumn{2}{c|}{pos. (2D)} & \multicolumn{3}{c|}{distance}                        \\ \hline
\end{tabular}
\vspace{1mm}
\caption{Structure of the feature vectors of nodes. The first two sections refer to the one-hot encodings that specify the node types and the frame they belong to. Positions (\textit{pos.}) are defined by 2D euclidean coordinates. Speeds are expressed using 3 dimensions for the linear and angular velocities in the plane. Orientations are given by the corresponding sine and cosine values. All metric values are in the robot's reference frame.}
\label{tbl:nodefeatures}
\end{table}

\par
\textbf{Edge} features were implemented differently for the experiments depending on the blocks used.
Some GNN blocks such as GAT or GCN, do not support edge features or labels, so no edge information is provided when they are used.
R-GCN blocks support edge labels, so a different label is used for each possible type of relation (\textit{e.g.}, human\nobreakdash-human, human\nobreakdash-room, wall\nobreakdash-room).
MPNN blocks treat edge information as features not limiting it to identifiers.
Therefore, besides containing values identifying the kind of relationship as a one-hot encoding, edge features also include the distance between the two entities being linked when using MPNN blocks.

\section{Experimental results}\label{sec:experiments}
Based on the assumption that in real life scenarios we can build on top of third party body trackers (\textit{e.g.}, \cite{rodriguezcriado2020multicamera,Qi2018}) and path planning systems, we proceed with the evaluation of the approach against the dataset presented in section~\ref{socnav2}.
Because all nodes are connected to their corresponding room node, the GNNs were trained to perform backpropagation based on the feature vector of the \textit{room} node in the last layer.
\par
Three GNN blocks were considered in the experiments: the two best-performing GNN blocks in~\cite{manso2020graph} (\textit{i.e.}, R-GCN~\cite{Schlichtkrull2018}, GAT~\cite{Velickovic2018}) and MPNN~\cite{gilmer2017neural}.
The implementations tested are based on the Deep Graph Learning library (DGL~\cite{wang2019deep}), using PyTorch~\cite{paszke2019pytorch} as backend.
\par
To benchmark the different architectures, \trainingSessions{} training sessions were launched using the SocNav2 dataset with a split of 47598 samples for training, 643 for evaluation and 643 for testing.
Given the variability of scenarios, 643 was considered a representative sample set size.
The hyperparameters were randomly sampled from the range values shown in Table~\ref{tbl:experiment_hp}.
Table~\ref{tbl:experiment_results} summarises the results obtained for the best model of each architecture, providing the performance on the different splits of the dataset.

\begin{table}[ht]
\centering
\begin{tabular}{|l|c|c|}
\hline
hyperparameter  & min  & max   \\ \hline
max. epochs     & \multicolumn{2}{c|}{1000}  \\ \hline
patience        & \multicolumn{2}{c|}{4}     \\ \hline
batch size      & 25   & 70     \\ \hline
hidden units    & 20   & 90     \\ \hline
\textit{attention heads} & 3    & 10     \\ \hline
\textit{number of bases} & 4    & 24     \\ \hline
learning rate   & 1e-4 & 5e-4   \\ \hline
weight decay    & 0.0  & 1e-6  \\ \hline
layers          & 3    & 9      \\ \hline
dropout         & 0.0  & 1e-6   \\ \hline
alpha           & 0.1  & 0.3    \\ \hline
\end{tabular}
\caption{Ranges of the hyperparameter values sampled. \textit{Attention heads} is only applicable to GAT blocks. \textit{Number of bases} is only applicable to R-GCN blocks.}
\label{tbl:experiment_hp}
\end{table}

\begin{table}[ht]
\centering
\begin{tabular}{|c|c|c|c|}
\hline
\multirow{2}{*}{GNN block}
        & training loss    & development loss & test loss       \\ 
        & (MSE)            & (MSE)            & (MSE)           \\
\hline
R-GCN   & 0.017347 & 0.040098   & 0.040607  \\
\hline
GAT     & 0.015188 & 0.037838   & 0.035818  \\
\hline
\textbf
{MPNN}  & 0.025020 & \textbf{\bestdevloss{}}   & \bestdevlossTEST{}  \\ \hline
\end{tabular}
\caption{The 3 GNN blocks tested along with their MSE for SocNav2.}
\label{tbl:experiment_results}
\end{table}

The training results obtained (see Table~\ref{tbl:experiment_results})
show that MPNN blocks delivered the best results, with a Mean Squared Error (MSE) of $\bestdevloss{}$ for the evaluation dataset.
The best model, which was selected based on the MSE on the evaluation split, yielded an MSE of $\bestdevlossTEST{}$ for the test split.
The best performing model was trained with a batch size of $57$, a learning rate of $2.5\text{e-}4$, weight decay regularisation of $1.0\text{e-}6$ and no dropout.
Its \textbf{network architecture} is a sequence of \textbf{6 MPNN blocks} with 40, 30, 21, 12 and 3 hidden units.
\par
\begin{figure*}[!ht] 
    \centering
    \subfloat[Fourteen humans.]{\includegraphics[width=0.45\textwidth,keepaspectratio=true]{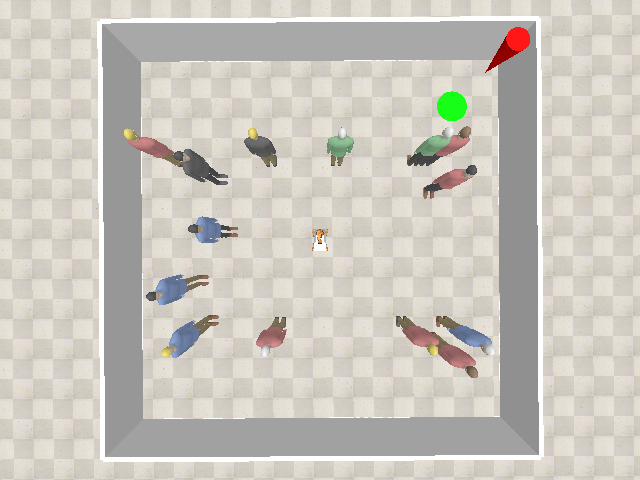} \label{fig:img_scn1_f0}}
    \hspace{0.4cm}
    \subfloat[Two humans.]{\includegraphics[width=0.45\textwidth,keepaspectratio=true]{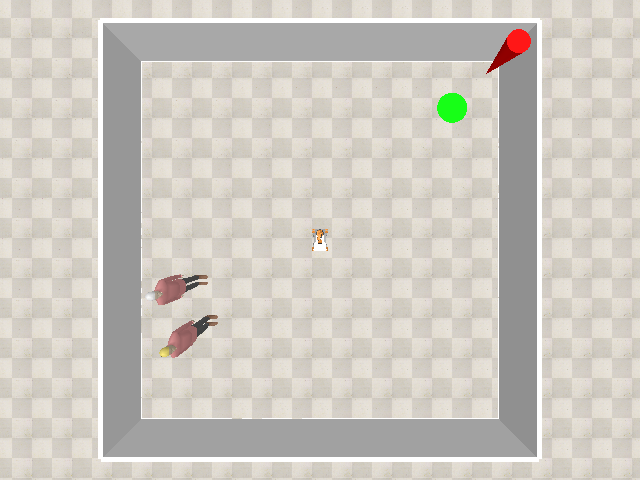} \label{fig:img_scn1_f3}}
    \caption{Two scenarios containing a different number of people. Results for these scenarios are shown in Fig.~\ref{fig:res1}.}
    \label{fig:scenario_1}
\end{figure*}
\begin{figure*}[!ht] 
    \centering
    \subfloat[First frame of the sequence.]{\includegraphics[width=0.45\textwidth,keepaspectratio=true]{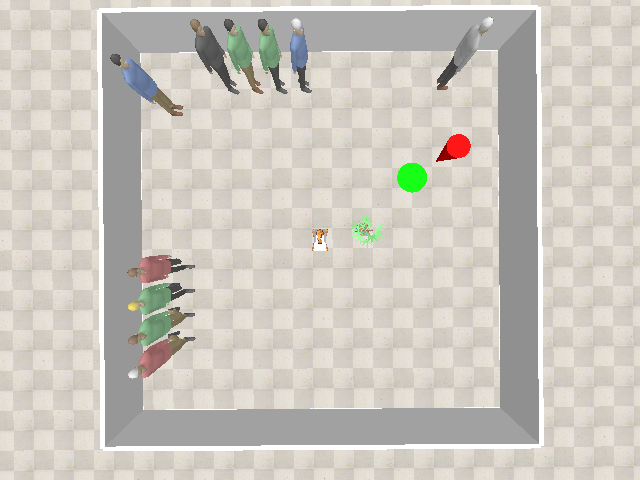} \label{fig:img_scn2_f0}}
    \hspace{0.4cm}
    \subfloat[Last frame of the sequence.]{\includegraphics[width=0.45\textwidth,keepaspectratio=true]{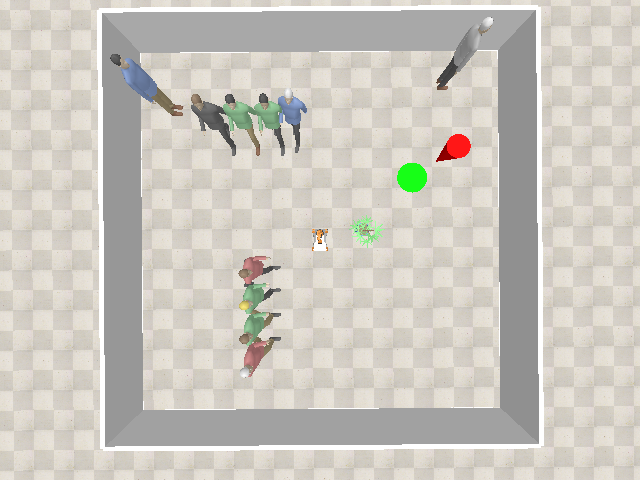} \label{fig:img_scn2_fN}}
    \caption{Scenario with two groups of people walking. Results for these scenarios are shown in Figs.~\ref{fig:res2_q1} and~\ref{fig:res2_q2}.}
    \label{fig:scenario_2}
\end{figure*}

To provide an intuition of the output of the network, the scenarios of Figs.~\ref{fig:scenario_1}, \ref{fig:scenario_2} and~\ref{fig:scenario_3} have been tested considering the output of the model for all the different positions of the robot in the room.
As a result, a heatmap representation of the network's response has been obtained for each tested scenario.
To ease the interpretation of each heatmap, the elements presented in the scenarios have been drawn over the image with the following representation: oriented blue circles for humans, small green circles for objects, a wider green circle for the goal position and red lines for interactions.
The horizontal and vertical axis of the room's frame of reference have also been depicted using black discontinuous lines to help distinguish the differences among the heat maps.

\par
\begin{figure*}[!ht] 
 \centering
\subfloat[Fourteen humans.]{\includegraphics[width=0.45\textwidth,keepaspectratio=true]{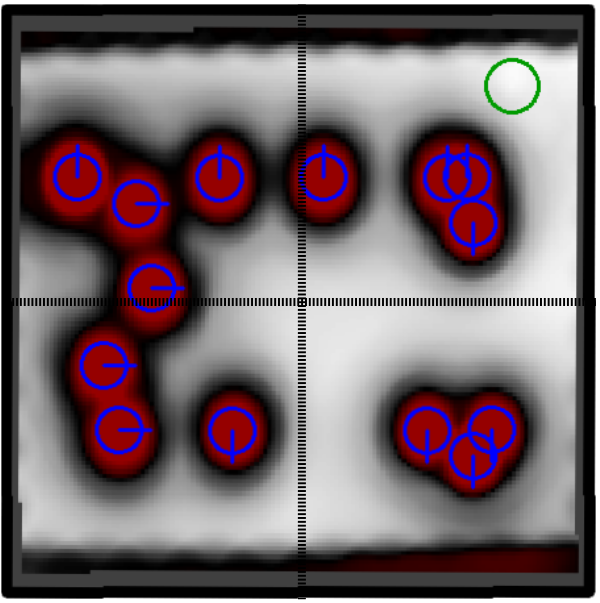} \label{fig:scn1_a}}
\subfloat[Two humans.]{ \includegraphics[width=0.45\textwidth,keepaspectratio=true]{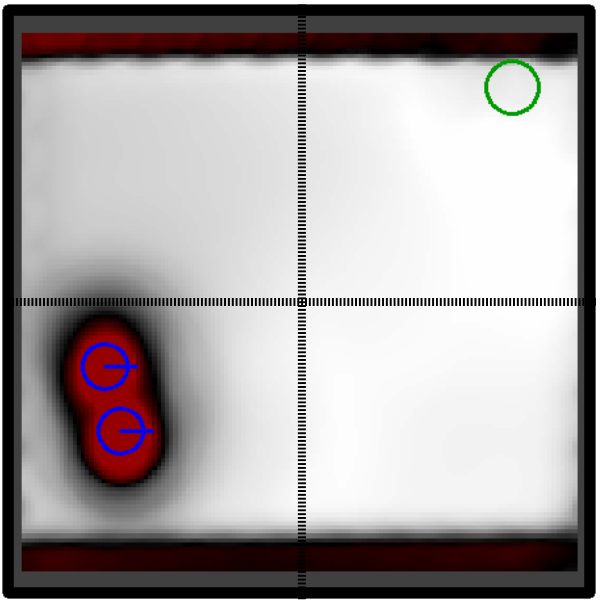}\label{fig:scn1_d}}
\caption{Output of the model for the two scenarios in Fig.~\ref{fig:scenario_1}. The response of the model is more strict for the case with fewer people.}
\label{fig:res1}
\end{figure*}

Fig.~\ref{fig:res1} shows the resulting generated maps for the first and last situations of the scenario of Fig.\ref{fig:scenario_1} considering the network output for Q1.
The different colours represent the output of the network.
A red colour is used to show a value near to 0 (unacceptable situation).
Grey tones represent the remaining range of values, where dark grey levels indicate lower values (high degree of discomfort) and a light one a high value (socially acceptable).
This test shows how the network adapts to differently populated environments.
For crowded spaces such as the one in Fig.~\ref{fig:img_scn1_f0}, the discomfort area of the humans narrows in relation to scenarios with less dense spaces.
For instance, the \textit{unacceptable} area of the humans in the bottom left of the room is wider in Fig.~\ref{fig:scn1_d} than in Fig.~\ref{fig:scn1_a}.
In addition, the response of the network increases in the positions near the walls if the number of people in the room is high (see the goal marked by a green circle in the right top corner of the images as a reference point).
This means that the positions near the limits of the room are considered more suitable for crowded environments.

\begin{figure*}[!ht] 
 \centering
\subfloat[Moving forward to the left.]{\includegraphics[width=0.333\textwidth,keepaspectratio=true]{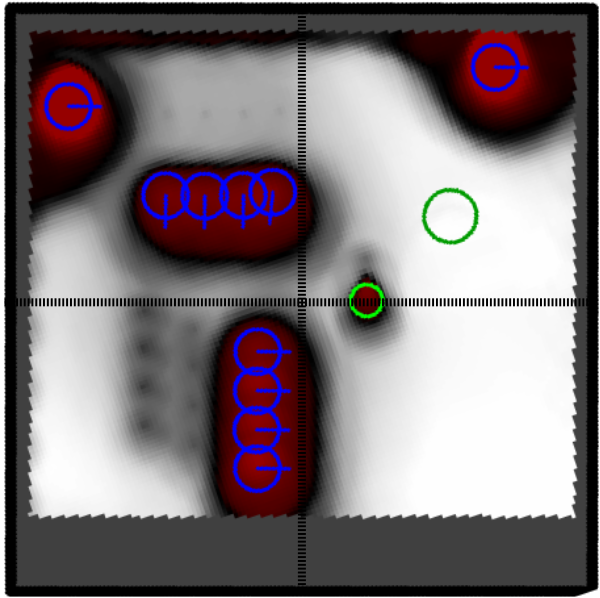} \label{fig:scn2_q1_a}}
\subfloat[Moving forward.]{ \includegraphics[width=0.333\textwidth,keepaspectratio=true]{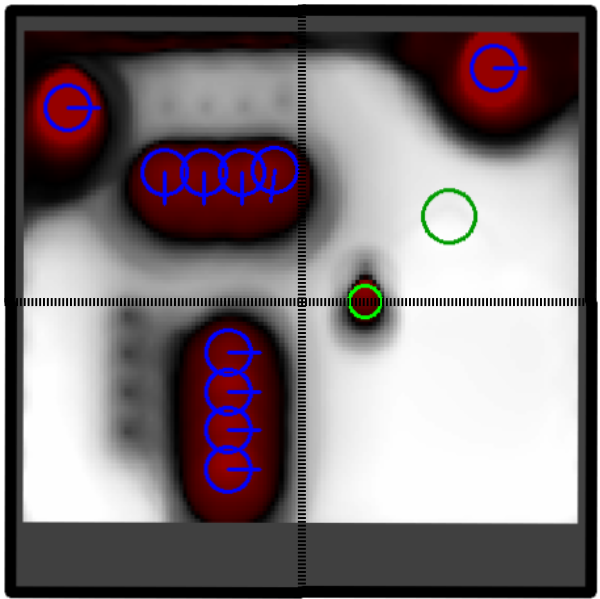}\label{fig:scn2_q1_b}}
\subfloat[Moving forward to the right.]{\includegraphics[width=0.333\textwidth,keepaspectratio=true]{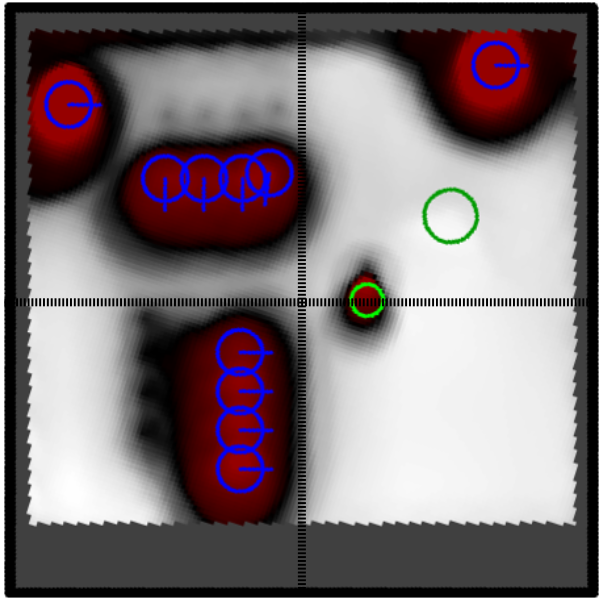} \label{fig:scn2_q1_c}} \\
\subfloat[Turning left.]{ \includegraphics[width=0.333\textwidth,keepaspectratio=true]{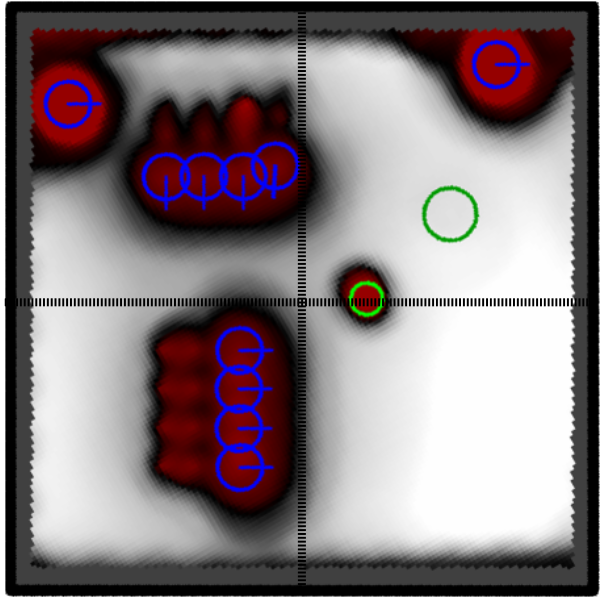}\label{fig:scn2_q1_d}}
\subfloat[Stopped.]{ \includegraphics[width=0.333\textwidth,keepaspectratio=true]{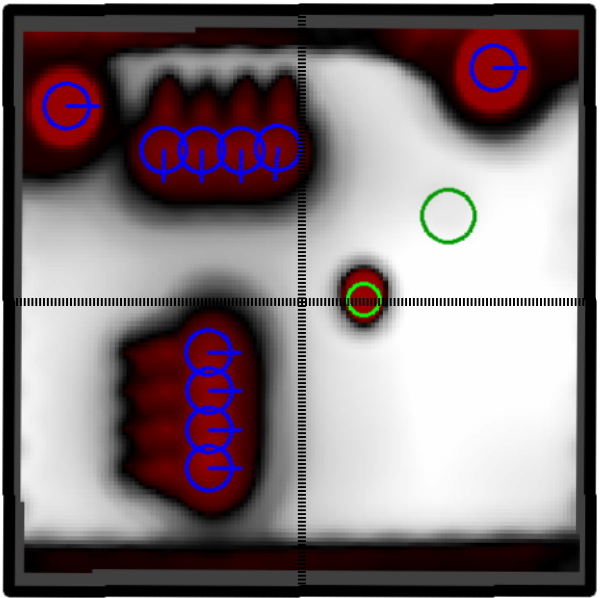}\label{fig:scn2_q1_e}}
\subfloat[Turning right.]{ \includegraphics[width=0.333\textwidth,keepaspectratio=true]{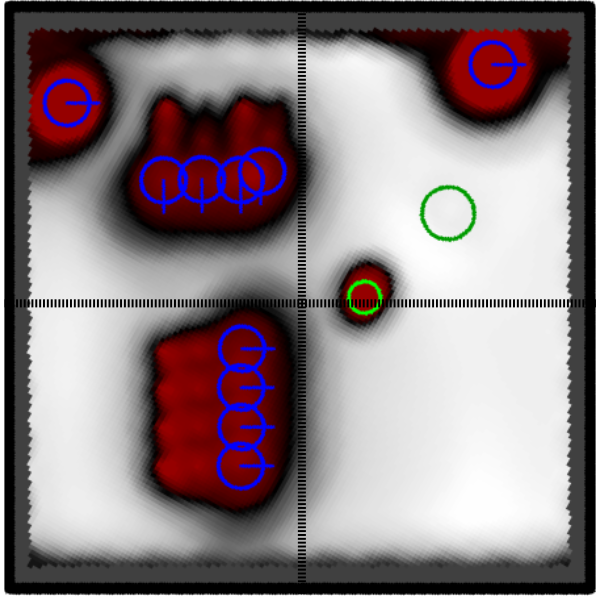}\label{fig:scn2_q1_f}}
\caption{Response of the model for Q1 for the scenario in Fig.~\ref{fig:scenario_2} considering different actions of the robot.}
\label{fig:res2_q1}
\end{figure*}

\begin{figure*}[!ht] 
 \centering
\subfloat[Moving forward to the left.]{\includegraphics[width=0.333\textwidth,keepaspectratio=true]{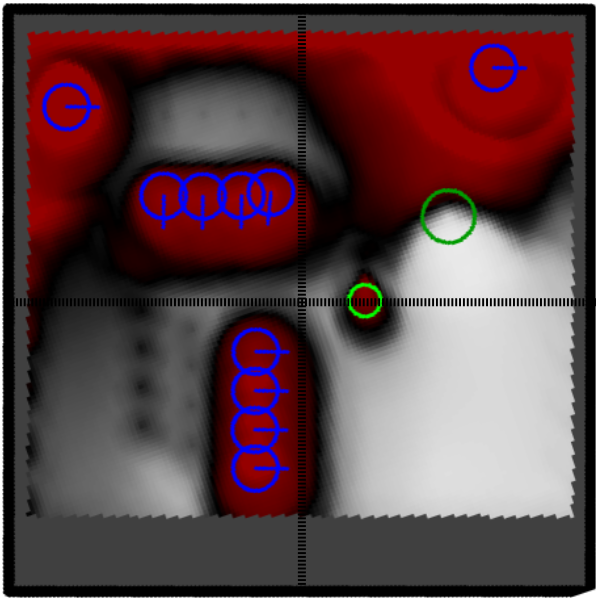} \label{fig:scn2_q2_a}}
 %\hspace{0.4cm}
\subfloat[Moving forward.]{ \includegraphics[width=0.333\textwidth,keepaspectratio=true]{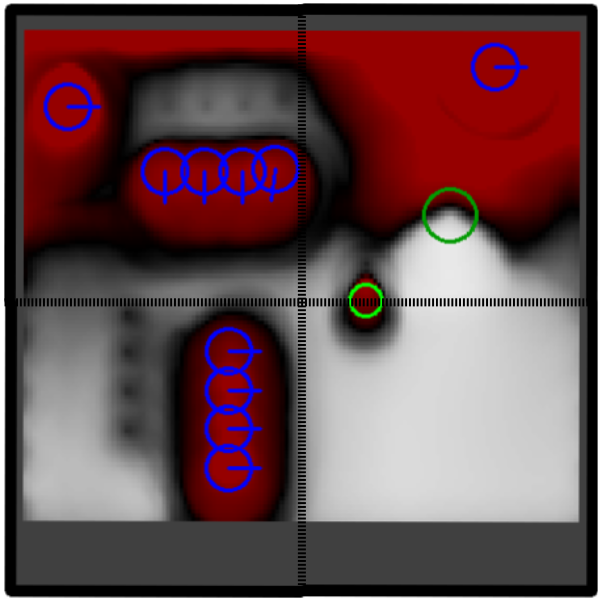}\label{fig:scn2_q2_b}}
 %\hspace{0.4cm}
\subfloat[Moving forward to the right.]{\includegraphics[width=0.333\textwidth,keepaspectratio=true]{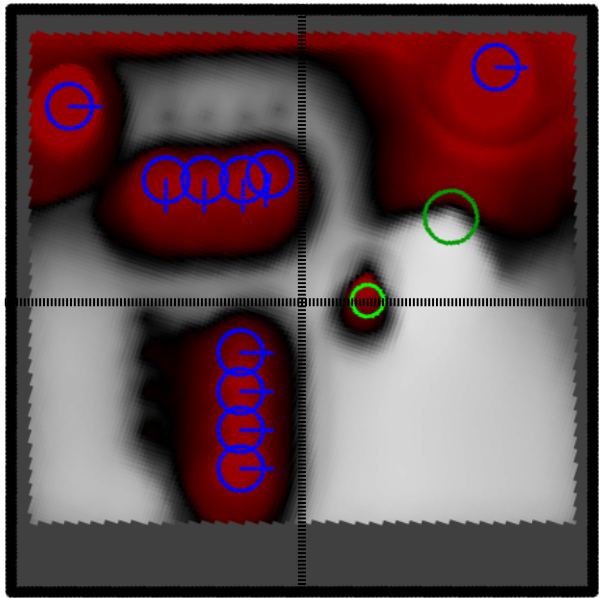} \label{fig:scn2_q2_c}} \\
\subfloat[Turning left.]{ \includegraphics[width=0.333\textwidth,keepaspectratio=true]{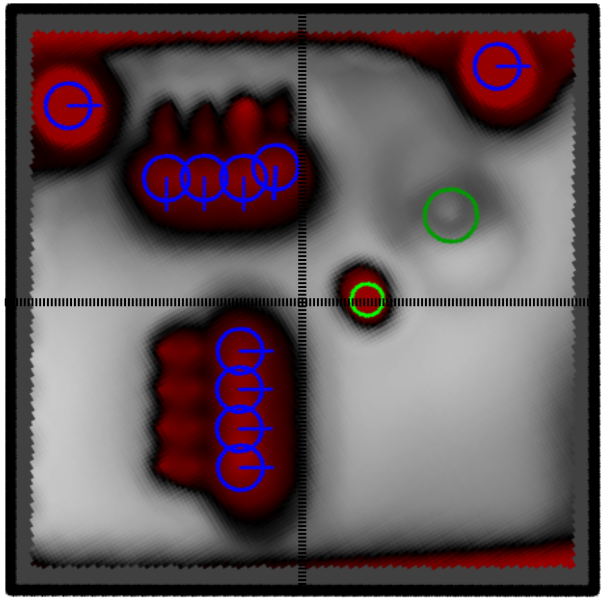}\label{fig:scn2_q2_d}}
 %\hspace{0.4cm}
\subfloat[Stopped.]{ \includegraphics[width=0.333\textwidth,keepaspectratio=true]{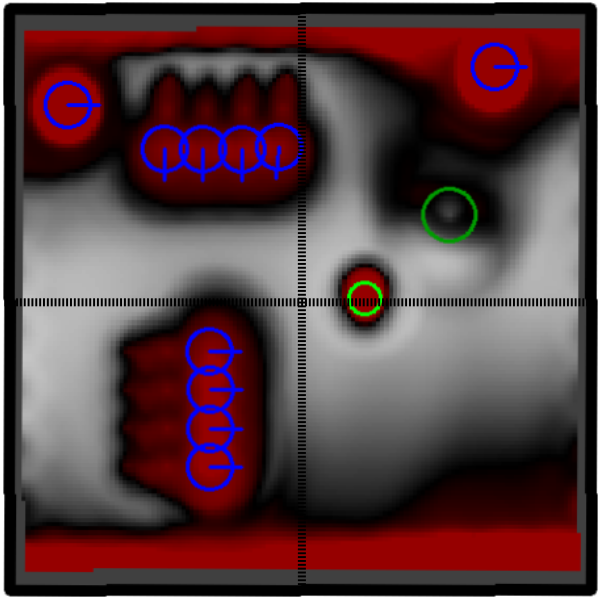}\label{fig:scn2_q2_e}}
 %\hspace{0.4cm}
\subfloat[Turning right.]{ \includegraphics[width=0.333\textwidth,keepaspectratio=true]{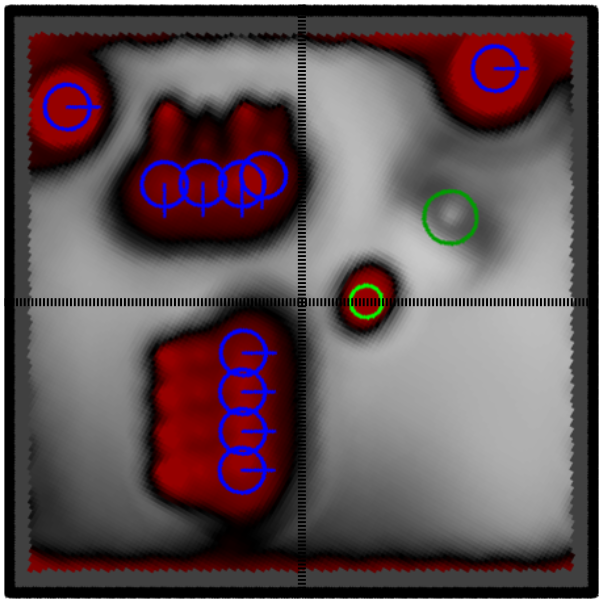}\label{fig:scn2_q2_f}}

\caption{Output of the model for Q2 for the scenario depicted in Fig.~\ref{fig:scenario_2} considering different actions of the robot.}
\label{fig:res2_q2}
\end{figure*}

The scenario in Fig.~\ref{fig:scenario_2} has been used to test how the actions of the robot have influence in the behaviour of the network.
Figs.~\ref{fig:res2_q1} and~\ref{fig:res2_q2} show the response of the network for Q1 and Q2, respectively.
From bottom to top, left to right, the actions of the robot for each image are the following: turning left, stopped, turning right, moving forward to the left, moving forward, moving forward to the right.
As shown in Fig.~\ref{fig:res2_q1}, the \textit{unacceptable} area (red area) of moving people changes according to their motion direction and the robot actions, while for standing humans such an area remains almost unalterable. 
In this way, when the robot moves forward (Fig.~\ref{fig:scn2_q1_b}) the red area of the group of people moving in the opposite direction extends towards the direction of the movement.
However, for the same action of the robot, the red area of the group of people moving in the horizontal direction keeps centered in the vertical-axis' position of the humans.
For this second group, the unacceptable area extends forwards or backwards when the robot moves to the left (figure \ref{fig:scn2_q1_a}) or to the right (figure \ref{fig:scn2_q1_c}).
When the robot is stopped or turning without translation, the positions with the lowest scores elongate towards the opposite direction of the movement of the humans (figures \ref{fig:scn2_q1_d}, \ref{fig:scn2_q1_e} and \ref{fig:scn2_q1_f}).
These positions correspond to the trajectory followed by the humans during the sequence, therefore the network response can be considered consistent with the situation.

\par
As expected, the response of the network regarding humans is maintained for Q2 (Fig.~\ref{fig:res2_q2}), but in this case the positions with low values increase according to the goal position and the action of the robot.
For instance, moving forward leaving the goal behind has a very low score.
Thus, when the goal is situated behind the robot, the best scoring actions are turning right or left according to the relative position of the goal.

\begin{figure*}[!ht] 
 \centering
\subfloat[With interactions.]{\includegraphics[width=0.45\textwidth,keepaspectratio=true]{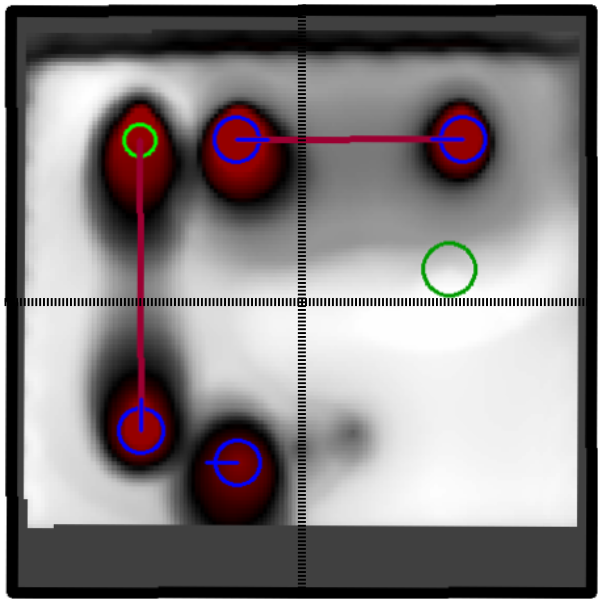} \label{fig:scn3_a}}
 \hspace{0.4cm}
\subfloat[Interactions removed.]{ \includegraphics[width=0.45\textwidth,keepaspectratio=true]{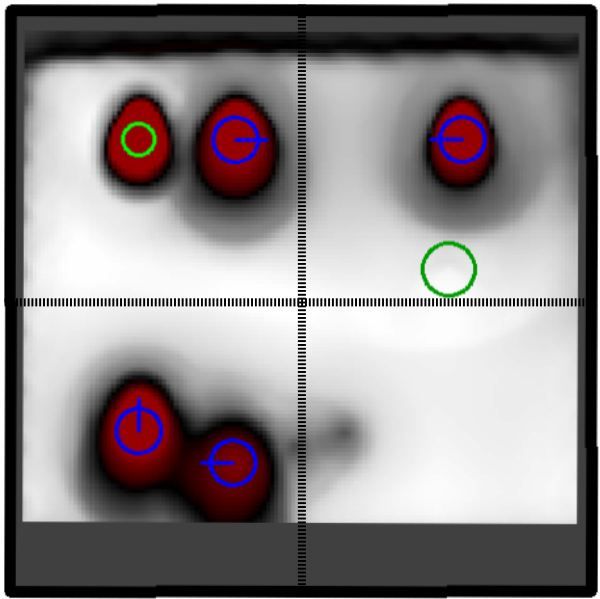}\label{fig:scn3_b}} 
\caption{The output of the model for the sequence depicted in Fig.~\ref{fig:scenario_3}. Fig.~\ref{fig:scn3_a} is the output of the model with the sequence in its original form. Fig.~\ref{fig:scn3_b} is the output of the model with the interactions removed. It is apparent that the response of the model for the perpendicular interaction is lower than that of the parallel one. This aligns with the intuition that the robot would be less disturbing if crossing perpendicularly than along the interaction line.}
\label{fig:res3}
\end{figure*}

To test the network response to potential interactions between humans or humans and objects, the scenario of Fig.~\ref{fig:scenario_3} has been used with the robot moving forward.
Results for this scenario with and without interactions for Q1 are shown in Fig.~\ref{fig:res3}.
As can be observed in Fig.~\ref{fig:scn3_a}, the interaction between the human an the object produces lower values than the interaction between the two humans.
This is consistent with the action of the robot, since the human-object interaction is taking place in the direction of the movement of the robot.
As a consequence, the interruption caused by the robot action is more intense than the one that is produced in the human-human interaction. 
If no interactions are taking place (Fig.~\ref{fig:scn3_b}), the areas between the two humans in the top of the image and the human and the object in the left are considered socially acceptable positions.
Thus, the network is properly generalising the different kinds of situations.
Another interesting result that can be seen in Fig.~\ref{fig:scn3_b} is the different treatment of humans an objects when objects are not being used by humans.
Specifically, being close to an object has a high response, while being close to a human
is not considered acceptable.

\par

\begin{figure}[ht]
\centering
\includegraphics[width=0.9\columnwidth]{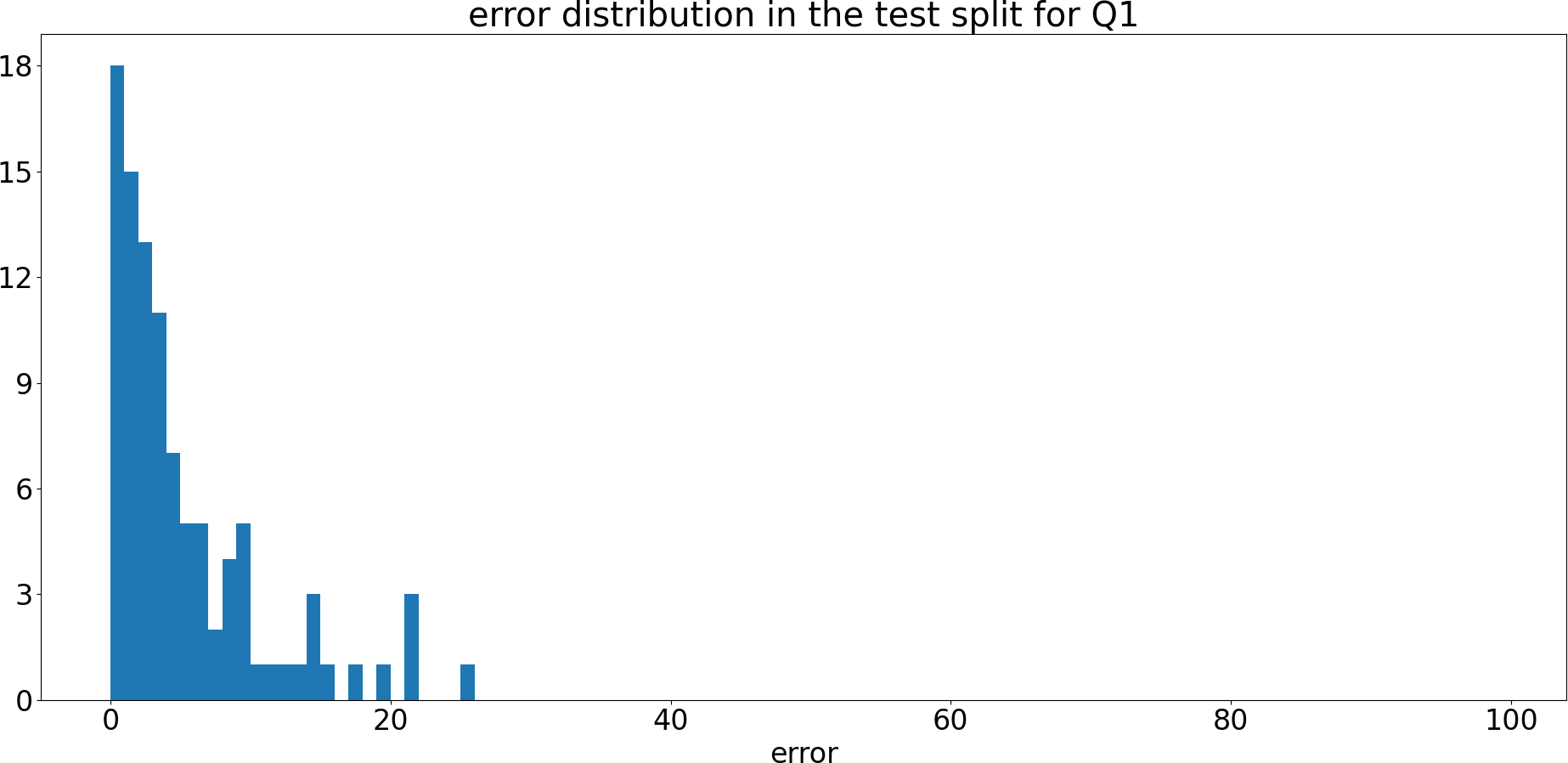}
\caption{Histogram of the absolute error in the test dataset for Q1.}
\label{fig:histogramQ1}
\end{figure}

\begin{figure}[ht]
\centering
\includegraphics[width=0.9\columnwidth]{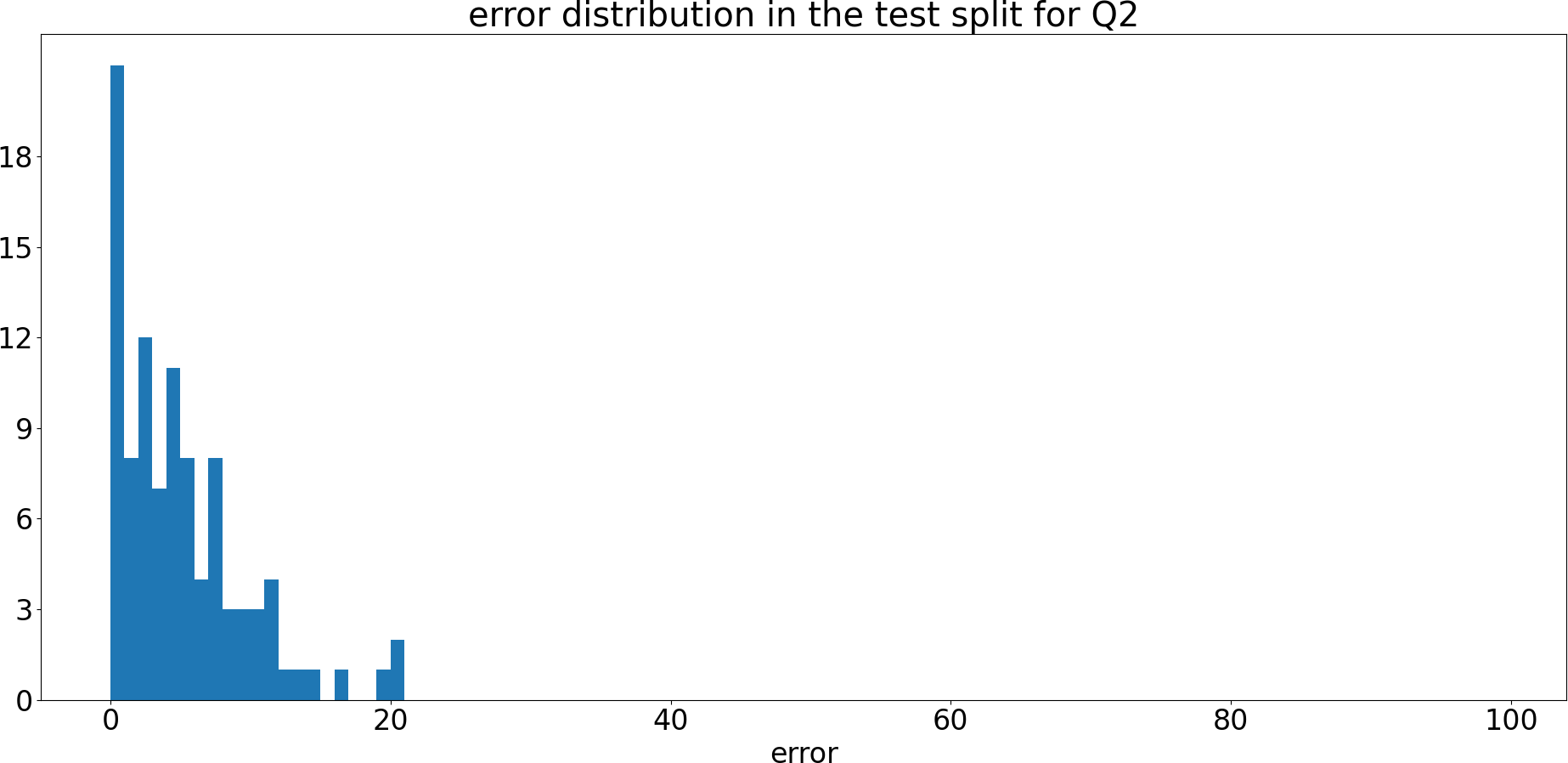}
\caption{Histogram of the absolute error in the test dataset for Q2.}
\label{fig:histogramQ2}
\end{figure}

Due to the subjective nature of the scores in the dataset (human feelings are utterly subjective), there is some level of disagreement even among humans.
To compare the performance of the network with human performance, we used a subset of the samples in SocNav2 which was labelled twice by each of the subjects (the same subset used to obtain Tables~\ref{tab:inter_intra_Q1} and~\ref{tab:inter_intra_Q2}).
Using the mean of the 8 scores that were provided for each scenario as a reference, the MSE for each of the participants was computed.
The average MSE was $\humanloss{}$, so we use that value as an indicative of human accuracy.
This means that the network performs close to human accuracy (even slightly better $\bestdevlossTEST{}$).
Figs.~\ref{fig:histogramQ1} and~\ref{fig:histogramQ2} show the histograms of the error of the model in the test split of the dataset for Q1 and Q2.
In~\cite{manso2020graph} we compared our results disregarding speed with~\cite{Vega2019} and achieved a considerably lower mean squared error (0.022 versus 0.12965).
Although the comparison was favourable, it is not entirely fair as the approaches have slightly different goals.
We are aware of other researchers currently working with the dataset used in this paper and SocNav1~\cite{manso2020socnav}, but there are no published works to compare with at the time of writing.

\par
\par

\section{Conclusions}
\label{sec:Conclusions}
Most approaches introduced in section~\ref{sec:Intro} deal with modelling human intimate, personal, social and interaction spaces instead of social inconvenience, which is a more general term.
To the best of our knowledge, all papers modelling discomfort around robots disregard information such as explicit interactions, trajectories or speed.
This paper tackled these issues with a specific scenario-to-graph transformation and a graph neural network architecture composed of 6 MPNN blocks.
\par
The results obtained are close to human accuracy and improve those in~\cite{manso2020graph} not only in terms of MSE but also in terms of the features considered (\textit{i.e.},~the trajectory and speed of the robot and the humans).
The results confirm that the discomfort is only skewed to the front of the humans when there is movement involved, which was initially hypothesised in~\cite{manso2020graph}.
The results also show that: \textbf{a)} the model adapts to a variable density of humans (see Fig.~\ref{fig:res1}); \textbf{b)} static humans are considered more carefully; and \textbf{c)} the model is able to consider the interactions which have been given explicitly.
\par
Future works point to user profiling and personalisation, as well as considering the activity of the humans and their gaze as done in works such as~\cite{Chen2020}.
Also, an ongoing line of research explores ways of linking the output of the GNN (questions Q1 and Q2) to driving control of the robot.
An end-to-end solution is a possibility but complicates the acquisition of labelled examples and the modulation of the final control action. An interesting alternative would be to use the output of the GNN as an additional restriction to be fulfilled by a Model Predictive Controller~\cite{Neunert2016}.
\par
The code to test the resulting GNN model, including the code implementing the scenario-to-graph transformation and the code to train the model suggested, has been published in a public repository as open-source software: \url{https://github.com/gnns4hri/sngnnv2}.

\bibliographystyle{spmpsci}      % mathematics and physical sciences
\bibliography{bibliography}   % name your BibTeX data base

\end{document}